\pdfoutput=1

\documentclass[11pt]{article}

\usepackage{emnlp2021}

\usepackage{times}
\usepackage{latexsym}
\usepackage{graphicx}
\usepackage{xurl}

\usepackage[T1]{fontenc}

\usepackage[utf8]{inputenc}

\usepackage{microtype}

\usepackage{booktabs}

\title{Small-Bench NLP: Benchmark for small single GPU trained models in Natural Language Processing}

\author{Kamal Raj Kanakarajan \and Bhuvana Kundumani \and Malaikannan Sankarasubbu\thanks{*Equal Contribution: Kamal and Bhuvana focussed on the coding and implementation under the guidance of Malaikannan.}\\
         SAAMA AI Research Lab, Chennai, India \\
         {\texttt\small \{kamal.raj, bhuvana.kundumani, malaikannan.sankarasubbu\}@saama.com}
         }

\begin{document}
\maketitle

\begin{abstract}

Recent progress in the Natural Language Processing domain has given us several State-of-the-Art (SOTA) pretrained models which can be finetuned for specific tasks. These large models with billions of parameters trained on numerous GPUs/TPUs over weeks are leading in the benchmark leaderboards. In this paper, we discuss the need for a benchmark for cost and time effective smaller models trained on a single GPU. This will enable researchers with resource constraints experiment with novel and innovative ideas on tokenization, pretraining tasks, architecture, fine tuning methods etc. We set up Small-Bench NLP, a benchmark for small efficient neural language models trained on a single GPU. Small-Bench NLP benchmark comprises of eight NLP tasks on the publicly available GLUE datasets and a leaderboard to track the progress of the community. Our ELECTRA-DeBERTa (15M parameters) small model architecture achieves an average score of \textbf{81.53} which is comparable to that of BERT-Base's 82.20 (110M parameters). Our models, code and leaderboard are available at  \url{https://github.com/smallbenchnlp}
\end{abstract}

\section{Introduction}

Recent research in machine learning and deep learning have delivered state-of-the-art results in various tasks across different domains. AlphaFold 
\footnote{\url{https://deepmind.com/blog/article/alphafold-a-solution-to-a-50-year-old-grand-challenge-in-biology}} from Deepmind – recognised as a solution to the 50-year old protein folding problem – has made a significant scientific breakthrough. This state-of-the-art attention based neural network model was trained on 16 TPUv3s (which is 128 TPUv3 cores or roughly equivalent to ~100-200 GPUs) for a few weeks. Exponential growth in the computing capabilities, availability of large datasets together with efficient and optimised algorithms have resulted in such highly accurate state-of-the-art models across different domains.  
\begin{figure*}[!ht]
\centering
\includegraphics[width=\textwidth,height=3.5cm]{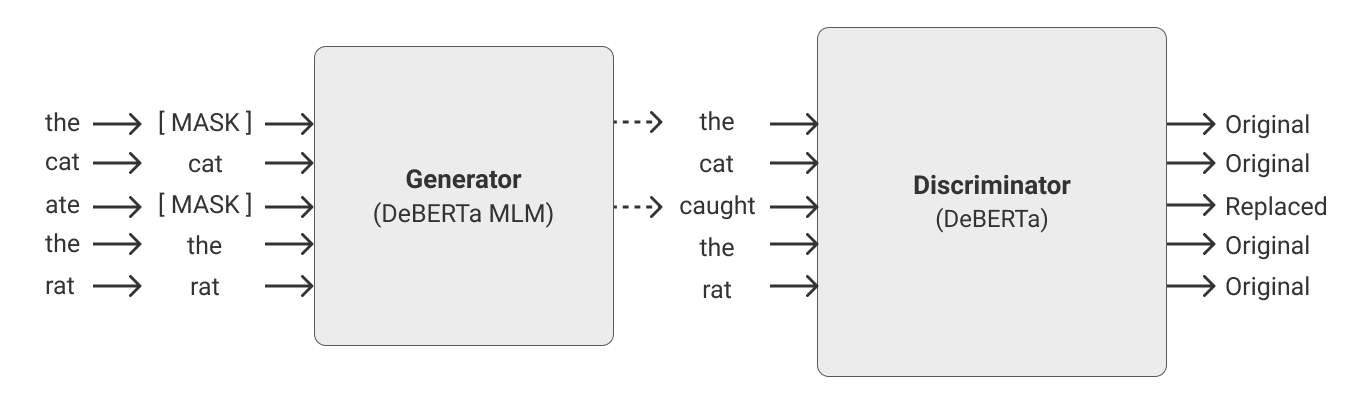}
\caption{ELECTRA-DeBERTa architecture}
\label{fig:ELECTRA-DeBERTa}
\end{figure*}

Similarly, language modelling task at the core of the natural language understanding has leveraged  unsupervised learning techniques together with availability of huge corpus of data to build contextualized dense vector representations for the words. Recent large-scale transformer-based neural language models like GPT-2 \citep{radford2019language, brown2020language}, BERT \citep{devlin2018bert}, RoBERTa \cite{liu2019roberta}, XLNet \cite{yang2020xlnet}, UniLM \cite{dong2019unified}, ELECTRA \cite{clark2020electra}, T5 \cite{2020t5}  have released pre-trained language models. The learned features from these pre-trained neural language models are then used for several supervised downstream tasks using the transfer learning approach. 

These state-of-the-art models have millions/billions of parameters and are time and compute-intensive. BERT-Large has 340M parameters, GPT-2 XLarge \citep{radford2019language} has 1.5B parameters, T5-11B \citep{2020t5} has 11B parameters and GPT-3 \citep{brown2020language} has 175B parameters. In the recent past, we can see that the parameters of such language models have gone from hundreds of millions to billions in number. Training such large models cost huge sums of money. This trend with institutions competing on the leaderboards with ever-increasing compute-intensive large billion parameter models has significantly increased the financial costs  and the carbon footprint. In addition to the financial and environmental costs, these large models as mentioned in \cite{bender2021dangers} also increase opportunity cost as researchers will be steered away from directions requiring lower resources. 

In order to foster research and to encourage experimenting with innovative and novel ideas in  tokenization, neural network architecture, pre-training tasks and finetuning methods for language models, we believe there has to be a benchmark for smaller neural language models with few million parameters that can be trained on a single GPU. This benchmark will enable researchers and practitioners from diverse backgrounds to experiment and validate their models built using relatively smaller computational resources. It will also provide an opportunity to make Natural Language Processing (NLP) research more inclusive and accessible to a large community. 

\paragraph{}In this paper,
\begin{enumerate}
    \item create Small-Bench NLP, a benchmark for evaluating the performance of smaller neural language models trained on a single GPU.
    \item we present our results of single GPU (V100-16GB of RAM) trained models with a smaller architecture on eight different annotated datasets part of the GLUE \citep{wang2019glue} benchmark.
    \item we show that our ELECTRA-DeBERTa small model (15M parameters), which has DeBERTa encoders with disentangled attention in ELECTRA's generator/discriminator architecture achieves an average score of \textbf{81.53} which is comparable to the average score 82.20 of BERT-Base (110M parameters).
\end{enumerate}

\section{Transformers and language modeling}

With the introduction of Transformers \citep{vaswani2017attention} for machine translation, many subsequent research work uses the encoder/decoder blocks of the transformers in their architecture. BERT \cite{devlin2018bert} uses the transformer's encoder architecture and introduced two new pre-training approaches - Masked language Modelling (MLM) and Next sentence prediction (NSP). With BERT model architecture fixed, RoBERTa \citep{liu2019roberta} uses larger Byte Pair Encoding and dynamic masking. RoBERTa also shows the performance of the model was comparable or even slightly better without the NSP task. Unlike BERT which uses sentence pairs as inputs, RoBERTa uses contiguous sentences from multiple documents as inputs. 

ELECTRA \citep{clark2020electra} introduces a sample-efficient replaced token prediction task in contrast to BERT’s MLM task. A small generator network produces plausible alternatives for the masked tokens and a discriminator network predicts whether the tokens are original or replaced. ELECTRA’s approach yields performance comparable to RoBERTa and XLNet \citep{yang2020xlnet} while using a quarter of their compute. It is interesting to note that the ELECTRA-small model with 14M parameters has better performance than GPT-2 Base model with 117M parameters and comparable performance to BERT-Base with 110M parameters.

DeBERTa \citep{he2021deberta} introduces a disentangled attention mechanism and an enhanced mask decoder for the transformer-based neural language model. These novel techniques have enabled DeBERTa model to surpass human performance on the SuperGLUE \citep{wang2020superglue} benchmark.

\section{Small-bench NLP}

Motivated by ELECTRA-small results which showed smaller parameter models with innovative techniques can achieve comparable performance, we set out to experiment with a combination of ELECTRA and DeBERTa architecture on a single GPU machine. Our model performance on GLUE datasets was very impressive. However, we did not find a benchmark for smaller models trained with resource constraints. To this end, we created Small-Bench-NLP, a benchmark for smaller models trained on a single GPU. Since running a few iterations on a single GPU is significantly cheaper (few hundreds of dollars), we feel that it will enable researchers to experiment with different techniques - whole word masking with replaced token detection, different attention mechanisms to name a few. 

\section{Experiments}

\subsection{Pretraining Data}
We pretrain all our models on the OpenWebTextCorpus\footnote{\url{https://skylion007.github.io/OpenWebTextCorpus/}}, an open-source effort to reproduce OpenAI’s WebText dataset. OpenWebTextCorpus totals 38GB of uncompressed text in the general domain. It has millions of webpages text from reddit URLs that have been deduplicated and filtered. Given the single GPU constraint, we choose OpenWebTextCorpus to take advantage of the large freetext corpus generated by a community over BookCorpus and Wikipedia corpus. 
\label{subsection:Pretraining data}

\subsection{Pretraining}
We pretrain our small models from scratch on BERT, RoBERTa, DeBERTa and ELECTRA-DeBERTa model architecture on OpenTextWebCorpus data mentioned in ~\ref{subsection:Pretraining data}. We have not pretrained or finetuned for the ELECTRA architecture. We present the results of the ELECTRA-small model trained on OpenTextWebCorpus data from the ELECTRA paper. Inspired by the replaced token detection task in ELECTRA, we experiment with a new architecture ELECTRA-DeBERTa. In ELECTRA-DeBERTa architecture, we replace the BERT transformers in ELECTRA (Masked Language Modeling and Replaced token Detection) with DeBERTa encoders. We do not make any other changes to the ELECTRA model. Refer Figure \ref{fig:ELECTRA-DeBERTa} for the model architecture. 

BERT, ELECTRA and ELECTRA-DeBERTa use a pair of sentences as inputs to the model. However, RoBERTa and DeBERTa use a single sentence sampled contiguously from the text as inputs to the model. 

All our model parameters are almost identical to ELECTRA-small model. The models have transformer encoders with 12 layers, 256 hidden size, 4 attention heads and Feed Forward Network (FFN) with inner hidden size of 1024. Model parameters range from 14M to 24M parameters. For all of the models, we use the default tokenizer and vocab size of the corresponding model architecture. All our models including BERT are trained without sentence contrastive task (Next Sentence Prediction) and with dynamic masking. 

All the Masked Language models (BERT, RoBERTa, DeBERTa) models are trained for 1.45M steps. ELECTRA and ELECTRA-DeBERTa are trained for 1M steps. All the models are pretrained with a maximum sequence length of 128 tokens with a batch size of 128. We use a peak learning rate of 5e-4 and Adam \cite{kingma2017adam} weight decay optimizer for all models. A learning rate scheduler with linear decay and warmup of 10000 steps is adopted. We pretrain all of our models on a single V100 16GB GPU.  

\begin{table*}[!ht]
\begin{center}
\resizebox{\textwidth}{!}{\begin{tabular}{lccccccccc}
\specialrule{1pt}{1.5pt}{1.5pt}
                & CoLA  & SST   & MRPC  & STS      & QQP   & MNLI  & QNLI  & RTE   & AVG   \\
Metric         & MCC   & Acc   & Acc   & Spearman & Acc   & Acc   & Acc   & Acc   &       \\
\midrule
BERT            & 45.00 & 90.14 & 86.27 & 84.46    & 88.59 & 79.58 & 87.22 & 65.70 & 78.37 \\
RoBERTa         & 44.72 & 89.45 & 85.30 & 84.02    & 89.84 & 79.51 & 87.39 & 66.42 & 78.33 \\
DeBERTa         & 47.82 & 90.36 & \textbf{88.49} & 84.62    & 88.31 & 78.11 & 86.67 & 67.87 & 79.03 \\
ELECTRA         & 56.80 & 88.30 & 87.40 & \textbf{86.80}    & 88.30 & 78.90 & 87.90 & 68.50 & 80.36 \\
ELECTRA-DeBERTa & \textbf{57.50} & \textbf{90.40} & 88.22 & 86.74    & \textbf{90.44} & \textbf{81.78} & \textbf{88.10} & \textbf{69.09} & \textbf{81.53} \\
\specialrule{1pt}{1.5pt}{1.5pt}
\end{tabular}}
\caption{Results of single GPU models finetuned on GLUE \cite{wang2019glue} benchmark datasets. (MCC - Matthew's Correlation Coefficient, Acc - Accuracy, Spearman Correlation Coefficient ). All models are small models unless specified explicitly.}
\label{tab:small-nlp}
\end{center}
\end{table*}

\subsection{Datasets}
To evaluate the performance of our models, we finetune our models on eight datasets in the General Language Understanding Evaluation (GLUE)\citep{wang2019glue} benchmark. GLUE consists of nine datasets based on six different NLP tasks - question answering, linguistic acceptability, sentiment analysis, text similarity, paraphrase detection and Natural Language Inference (NLI). We train our models on the train set and evaluate our models on the publicly available development sets. We do not evaluate on the WNLI dataset due to the issues on the dev and test distributions. \footnote{\url{https://gluebenchmark.com/faq}}. Detailed description of the datasets are available in the \citep{wang2019glue}
\label{section:datasets}

\subsection{Finetuning}
We finetune the pretrained models on BERT, RoBERTa, DeBERTa and ELECTRA-DeBERTa small architecture on the datasets detailed in ~\ref{section:datasets}. All the models are finetuned from the respective pretrained checkpoints and we do not apply any standard tricks used by the GLUE leaderboard submissions like XLNet \cite{yang2020xlnet}, ALBERT \cite{lan2020albert}, etc.

For finetuning, we use Adam \cite{kingma2017adam} optimizer and a learning rate scheduler with warmup of ( 10\% of steps) followed by linear decay (90\% of steps). For BERT, RoBERTa and DeBERTa, we experiment with a learning rate between 1e-5 to 5e-5. For ELECTRA and ELECTRA DeBERTa we use a layer-wise learning-rate decay out of [0.9, 0.8, 0.7] and a learning rate between 1e-4 to 3e-4. We experiment with batch size [16, 32] and epochs [3,10]. We run 5-10 fine tuning runs on these datasets and we report the best score on the development set as the final score for the evaluation metric.

\section{Benchmark Results} 
We evaluate the datasets with the same evaluation metric mentioned in the GLUE \cite{wang2019glue} benchmark. The results of our models for the eight GLUE \cite{wang2019glue} datasets are shown in table \ref{tab:small-nlp}. We see that our experimental ELECTRA-DeBERTa architecture has significant gains over the BERT models (BERT, RoBERTa and DeBERTa) due to the Replaced Token Detection task of ELECTRA and  disentangled attention mechanism of DeBERTa. The scores for ELECTRA and ELECTRA-DeBERTa architecture are comparable to each other. It is very evident that the disentangled attention mechanism has given a 1 point increase in scores for the ELECTRA-DeBERTa architecture over the ELECTRA architecture.

\begin{table}[!ht]
\resizebox{\columnwidth}{!}{\begin{tabular}{lcc}
\specialrule{1pt}{1.15pt}{1.15pt}
Model           & Params & GLUE (Average) \\
\midrule
BERT            & 14M    & 78.37          \\
RoBERTa         & 22M    & 78.33          \\
DeBERTa         & 24M    & 79.03          \\
ELECTRA         & 14M    & 80.36          \\
ELECTRA-DeBERTa & 15M    & \textbf{81.53}          \\
\midrule
BERT-Base       & 110M   & \textbf{82.20} \\  
\specialrule{1pt}{1.15pt}{1.15pt}
\end{tabular}}
\caption{ Comparison of small models on the GLUE \cite{wang2019glue} benchmark dev set with BERT-Base model.}
\label{tab:compare-nlp}
\end{table}  

Further, we compare the performance of these single GPU trained models with that of the BERT-Base model in table \ref{tab:compare-nlp}. The BERT-Base results are taken from the ELECTRA paper. With all models pretrained on the same OpenWebText corpus for approximately similar number of steps, we attribute the gains in the model performance to the pretraining task and improved attention mechanism. We also observe that models with approximately 12\% to 20\% of the model parameters compared to that of the BERT Base (110M parameters) have an average score that is approximately ~1.8 to ~ 3.8 points lesser than that of BERT-Base average score of 82.20. Our ELECTRA-DeBERTa architecture achieves an average score of \textbf{81.53} which is comparable to the score 82.20 of the BERT-Base model.

\section{Conclusion and Future Work}
In this paper, we discuss the need for a benchmark for smaller language models trained on a single GPU. We present our results and show that smaller models trained with a few million parameters for few days have comparable performance to BERT-Base model with 110M parameters. To facilitate research and make it more accessible and inclusive to the community, we introduce the Small-Bench NLP for a variety of NLP tasks. We also set up a leaderboard to enable researchers and practitioners to validate their models. Our model trained using a Generator discriminator architecture (like ELECTRA) with DeBERTa's encoder which has the novel disentangled attention achieves an average score of \textbf{81.53} (15M parameters) which is comparable to the score 82.20 of BERT-Base (110M parameters).

We are interested in training and adding the scores for the other publicly available language models like XLNet \cite{yang2020xlnet}, MPNet \cite{song2020mpnet}, etc. We intend to work with collaborators and open-source contributors on extending small models trained on single GPU benchmark to multilingual/ other domains datasets (finance, biomedical etc). 

\section*{Acknowledgement}
We extend our sincere thanks to Samuel Gurudas for creating the diagrams in this research paper.

\bibliography{smallbenchnlp}
\bibliographystyle{acl_natbib}

\end{document}